\pdfoutput=1

\documentclass{article} % For LaTeX2e
\usepackage{latex_template,times}
\usepackage{hyperref}
\usepackage{url}
\usepackage{xcolor}
\usepackage{listings}
\usepackage{amssymb}
\usepackage{amsmath}
\usepackage{biblatex}
\usepackage[justification=centering]{caption}
\usepackage{mathtools}
\usepackage[many]{tcolorbox}
\usepackage{color,soul}
\usepackage{pdflscape}
\usepackage{booktabs}
\usepackage{mathtools}
\usepackage[flushleft]{threeparttable}
\usepackage[section]{placeins}

\usepackage[framemethod=tikz]{mdframed}
\usepackage[ruled,vlined]{algorithm2e}

\title{Analysis of classifiers robust to noisy labels}

\nipsfinalcopy % Uncomment for camera-ready version

\begin{document}

\maketitle
\author{
\begin{tabular}[t]{@{\extracolsep{14em}}*2c} 
\textbf{Alejandro Díaz} & \textbf{Damian Steele}\\
adia2600@uni.sydney.edu.au & dste5943@uni.sydney.edu.au
\end{tabular}
}

\begin{abstract}

We explore contemporary robust classification algorithms for overcoming class-dependant labelling noise: Forward, Importance Re-weighting and T-revision. The classifiers are trained and evaluated on class-conditional random label noise data while the final test data is clean. 
We demonstrate methods for estimating the transition matrix in order to obtain better classifier performance when working with noisy data. We apply deep learning to three data-sets and derive an end-to-end analysis with unknown noise on the CIFAR data-set from scratch. The effectiveness and robustness of the classifiers are analysed, and we compare and contrast the results of each experiment are using top-1 accuracy as our criterion.

\end{abstract}

\section{Introduction}

The advances in deep learning techniques, particularly in applications such as image classification, has put even greater importance on the accuracy of labels as we forge ahead in the big data era. The need for human input and intervention to identify, categorise or otherwise add context to data via labels is just one way that can also give rise to error and inaccuracies which we loosely term noise \cite{sukhbaatar2014training}. A variety of techniques have been proposed to tackle this problem and at the same time bring a number of assumptions of practical importance to machine learning practitioners \cite{algan2020image}.

Label noise is a significant obstacle when working with the massive are varied data-sets typically used in training advanced machine learning models, such as deep neural networks. Research has also shown that precision of the learned classifiers can be profoundly influenced by label noise \cite{1frenay2013classification, sukhbaatar2014training, 3zhang2016understanding}. Noisy labelled data-set training induces output loss because deep neural networks (DNNs) will easily over-fit the noise labels \cite{li2019learning}. For a basic cause, the dilemma is pervasive: manual technical labelling of each case on a wide scale is not possible, and researchers often turn to inexpensive yet incomplete surrogates \cite{5ergus2010learning, 35schroff2010harvesting}.

We set out to examine robust multi-class classification methods under a number of scenarios in which we leverage noisy labelled data-sets to derive insights and knowledge that applies to the underlying clean labelled data distribution. We develop two classifiers using known class-dependent, asymmetric flip rates which indicate the likelihood a member of each class has had its label changed. We then develop a procedure to estimate unknown flip rates applied to the CIFAR dataset. Finally, we demonstrate the effectiveness of our methods using top-1 accuracy from samples drawn from the true clean distribution.

% Many public data sources exist on the internet, but they appear to contain misleading labels. 

% The human-level output is dependent on vast training data with high-quality manual annotations, which are costly and time-consuming to obtain, considering the success of deep neural networks ( DNNs) in image classification tasks. 

% Therefore, we explore T-revision technique. All those approaches have been documented to perform very well empirically. 
% However, discrepancies are not guaranteed to disappear between the learned classifiers and the best ones for clean results, i.e. no statistical accuracy has been guaranteed.

% \hl{To learn and estimate the transition matrices effectively, we explore a transition-revision (T-Revision) strategy, leading to improved classifiers} \cite{xia2019t_revision}.

% \hl{In overall, we build a risk-consistent estimator based on deep learning to reliably tune the transition matrix. Specifically, by using instances that are close to anchor points, including those with large approximate posterior probabilities of the noisy class, we initialise the transition matrix first. Then, by adding a slack vector, we adjust the initial matrix, which will be studied and validated along with the classifier using only noisy data.}

\pagebreak

\section{Previous Work}

% We explore noisy channel aka loss-correction \cite{algan2020image}
% Random noise vs. class-dependant noise
% \hl{Importance Reweighting, symmetrical loss, forward, backward}
% vs. risk/classifier consistent algorithms.

Methods for label-noise learning can be generally divided into two categories: statistically consistent/inconsistent, and risk consistent classifiers. The first category looks to limit the impact noisy data through typically heuristic means i.e. expert selection of samples or label correction \cite{xia2019t_revision}. Methods belonging to the risk category assume label noise is randomly conditioned on the true labels \cite{sukhbaatar2014learning} and look to minimise (\ref{eq:risk-eq}) through a process of \textit{loss-correction} where $Q(f_{\theta}(x_i)) = p(\tilde{y}|f_{\theta}(x_i))$. $Q$ can be formulated with a noise transition matrix $T$ so that $Q(f_{\theta}(x_i)) = Tf_{\theta}(x_i)$ where each element of the matrix represents the transition probability of $\tilde{y}$ noisy to $y$ true, $T_{ij} = p(\tilde{y}=j|y=i)$  \cite{algan2020image}.  

\begin{equation}
    \hat{R}_{l,D}(f) = \frac{1}{N}\sum_i^N l(Q(f_{\theta}(x_i)),\hat{y_{i}})
\label{eq:risk-eq}
\end{equation}

% classification tasks in the face of noisy or corrupted label information. These approaches look

The noise transition matrix, denoting the probability of clean labels flipping into noisy labels, plays a central role in constructing statistically accurate classifiers in label-noise research. Current theories have demonstrated that by exploiting anchor points, the transition matrix can be trained (i.e. data points that almost definitely belong to a certain class). Given an instance x, if $P(Y=i|X=x) \approx 1$ and otherwise $P(Y=k|X=x) = 0$ where $k\neq i$ we have (\ref{eq:t-matrix}). However, the transition matrix will be improperly learned where there are no anchor points, and these formerly stable classifiers will greatly degenerate \cite{xia2019t_revision}.

% Scientific models have demonstrated that by exploiting anchor points, the transition matrix can be trained (i.e. data points that almost definitely belong to a certain class). However the transition matrix will be improperly learned where there are no anchor points, and certain formerly consistent classifiers will degenerate dramatically \cite{xia2019t_revision}.

% \begin{mdframed}
% \textbf{Anchor points:} Consistent algorithms' achievements rely on solid bridges, i.e. transition matrices that have been correctly trained. The definition of the anchor point was proposed in order to learn transition matrices \cite{22liu2015classification, 35scott2015rate}. 
% In the clean data domain, anchor points are described, i.e., an example x is an anchor point for class $i$ if $P(Y = i|X = x)$ is equivalent to one or near to one. Provided an $x$, if we have $P(Y = i|X = x) = 1$, for $k  \neq i, P(Y = k|X = x) = 0$. We have, then
% \end{mdframed}

\begin{equation}
    P( \overline{Y} = j|X = x) =  \sum_1^C  T_{kj}P(Y = k|X = x) = T_{ij} 
\label{eq:t-matrix}
\end{equation}

% \begin{figure}[h]
% \centering
% \includegraphics[width=12cm]{images/T-revision.png}
% \centering
% \caption{A more precise classifier can learn the T-revision approach when the transition matrix is improved}
% \label{fig:partial_noise}
% \end{figure}

\begin{figure}[h]
\centering
\includegraphics[width=9cm]{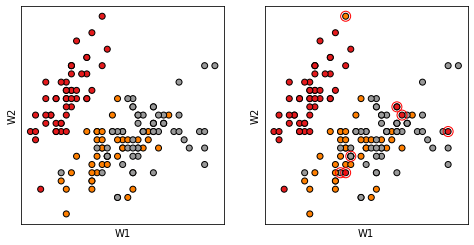}
\caption{Illustration of label noise. Left: True labels drawn from $D$, Right: Noisy labels drawn from $\hat{D}$. Red circles indicate corrupted or 'flipped' labels}
\label{fig:flip-rates}
\end{figure}

Sample Importance Weighting allows training to be made more effective by assigning weights to instances according to their estimated noisiness level. These methods look to minimise the empirical risk using a form (\ref{eq:reweighting-eq}) whereby a dynamic function $\beta(X,Y)$ determines the instance dependant weight \cite{algan2020image}. It is expected that a correctly labelled example has a large $\beta(X, Y)$ and contributes more to the risk, while an incorrectly labelled example has a smaller value and will contribute less \cite{wang2017multiclass}. However, these methods can also sometimes become biased towards a certain subset of data \cite{algan2020image}. In Figure \ref{fig:flip-rates}, we can see a number of 'flipped' class labels on the right-hand side. The estimated distributions of $D$ and $\hat{D}$ can be used to suppress the influence of these incorrect labels \cite{liu2015classification, 42xiao2015learning, wang2017multiclass, xia2019t_revision}.

\begin{equation}
    \hat{R}_{l,D}(f) = \frac{1}{N}\sum_{i=1}^N \beta(x_i,y_i)l(f_{\theta}(x_i),\hat{y_{i}})
\label{eq:reweighting-eq}
\end{equation}

% \section{Methods}

% REPORT STRUCTURE

% %Importance Reweighting
% \subsection{Pre-processing}
% \subsection{Formulation}
% \subsection{Loss Function}

% \section{Noise rate estimation method}
% \subsection{Pre-processing}
% \subsection{Formulation}
% \subsubsection{Loss Function}
% \subsubsection{Optimization}

% \section{Label noise methods with unknown flip rates}
% %T-Revision
% \subsection{Pre-processing}
% \subsection{Formulation}
% \subsection{Loss Function}

% \section{Experiments}

% \section{Discussion}

\section{Label Noise Methods with Known Flip Rates}

We examine robust methods for multi-class image classification using already known transition matrices and compare each method using top-1 accuracy. We examine the "Forward" method and the Importance of Re-weighting method. We introduce each formulation and the theoretical basis for robustness as well as details we employ to leverage these methods.  Experiments and results for each method are discussed at length in Section \ref{experiment-and-results}.

% , and contrast these methods' performance using the "Forward" method.

\subsection{Forward Learning}

The Forward Learning method incorporates the known T matrix into the learning procedure with the use of a noise adaption layer. Using the known noisy labelled dataset, we could simply train a network to best match the noisy labels $\hat{P}(\bar{Y}|X)$. Instead, we explicitly introduce the dependency on $T$, which allows us to compare the noisy labels to averaged noisy predictions corrupted by $T$ \cite{sukhbaatar2014learning, patrini2017making}. We can use the cross-entropy loss function (\ref{eq:forward-cross-entropy-1},\ref{eq:forward-cross-entropy-2},\ref{eq:forward-cross-entropy-3}) to illustrate this process, which consequently allows us to approximate $P(Y|X)$ . The noise layer serves as a normal linear layer but has no bias and its weights will range between 0 and 1 because the represent conditional probabilities $T_{ij}$ \cite{sukhbaatar2014learning}.

% where without $T$ would otherwise be a predictor of noisy labels $\hat{P}(\bar{Y}|X)$

\begin{equation}
\begin{array}{l}
\ell\left(\boldsymbol{e}^{i}, \hat{p}(\boldsymbol{y} \mid \boldsymbol{x})\right)=-\log \hat{p}\left(\tilde{\boldsymbol{y}}=\boldsymbol{e}^{i} \mid \boldsymbol{x}\right) \\
\end{array}
\label{eq:forward-cross-entropy-1}
\end{equation}

\begin{equation}
=-\log \sum_{j=1}^{c} p\left(\tilde{\boldsymbol{y}}=\boldsymbol{e}^{i} \mid \boldsymbol{y}=\boldsymbol{e}^{j}\right) \hat{p}\left(\boldsymbol{y}=\boldsymbol{e}^{j} \mid \boldsymbol{x}\right)
\label{eq:forward-cross-entropy-2}
\end{equation}

\begin{equation}
=-\log \sum_{j=1}^{c} T_{j i} \hat{p}\left(\boldsymbol{y}=\boldsymbol{e}^{j} \mid \boldsymbol{x}\right)
\label{eq:forward-cross-entropy-3}
\end{equation}

\subsection{Formulation}

Provided the noise matrix $T$ is non singular, a proper composite loss (for example, cross entropy) has the forward loss correction defined as (\ref{eq:forward-loss-correction}) \cite{patrini2017making}.

\begin{equation}
\ell_{\psi}(\boldsymbol{h}(\boldsymbol{x}))=\ell\left(T^{\top} \boldsymbol{\psi}^{-1}(\boldsymbol{h}(\boldsymbol{x}))\right)
\label{eq:forward-loss-correction}
\end{equation}

\begin{figure}[h]
\centering
\includegraphics[width=9cm]{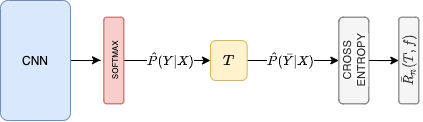}
\caption{Forward classifier.}
\label{fig:forward-classifier}
\end{figure}

The minimizer of the corrected loss under a noisy distribution is the same as the minimizer of the initial loss under that clean distribution (\ref{eq:forward-loss-minimizer}) \cite{patrini2017making}. Since $T$ is already known, we can minimise w.r.t the clean data distribution on the basis $\hat{P}(Y|X)$ approximates $P(Y|X)$. We make use of a custom neural network architecture in Figure \ref{fig:global_cnn} to iteratively learn $\hat{P}(Y|X)$ using the classifier in Figure \ref{fig:forward-classifier}.

\begin{equation}
\underset{h}{\operatorname{argmin}} \mathbb{E}_{\boldsymbol{x}, \tilde{y}} \ell_{\psi}^{\rightarrow}(\boldsymbol{y}, \boldsymbol{h}(\boldsymbol{x}))=\underset{\boldsymbol{h}}{\operatorname{argmin}} \mathbb{E}_{\boldsymbol{x}, \boldsymbol{y}} \ell_{\psi}(\boldsymbol{y}, \boldsymbol{h}(\boldsymbol{x}))
\label{eq:forward-loss-minimizer}
\end{equation}

% Since $T$ is already known, we can minimise w.r.t the clean data distribution on the basis $\hat{P}(Y|X)$ approximates $P(Y|X)$.  

\newpage

\subsection{Importance Re-weighting}

% Therefore, the T-revision approach provided significantly better transition matrix which results in the more precise classifier.

% \begin{algorithm}[H]
% \SetAlgoLined
% \SetKwInput{KwInput}{Input}
% \SetKwInput{KwFirst}{Stage 1}
% \SetKwInput{KwSecond}{Stage 2}
% \KwInput{Noisy training sample $D_t$; Noisy validation set $D_v$ }
% \KwFirst{Learn $\hat{T}$ }
%  1: Minimize the unweighted loss to learn $\hat{P}(\hat{Y}=i|X=x)$ without a noise adaption layer\;
%  2:Initialize $\hat{T}$ according to Eq. (\ref{eq:t-matrix}) by using instances with the highest $\hat{P}(\hat{Y}=i|X=x)$ as anchor points for the $i$-th class\;
% \KwSecond{Learn the classifier $f$ and $\triangle{T}$}
%  \caption{Reweight T-Revision Algorithm}
% \end{algorithm}

Sample importance weighting looks to adjust the effects of likely noisy labels using weights. The Importance Re-weighting method uses principles from domain adaptation which takes knowledge of the source domain (the noisy distribution) $\bar{D}$ to improve model performance in a target domain (the true or 'clean' distribution) $D$. Here the method uses the joint probability of $(X,Y)$ under the two distributions $D$ and $\bar{D}$ \cite{wang2017multiclass}.

% Wang et al. (2018) propose a multi-class classification with Importance Reweighting technique that uses the class posterior probabilities of a corrupted noisy domain where $\beta(X,Y) = P_{D}(X,Y)/P_{D_n}(X,\bar{Y})$ \cite{wang2017multiclass}.

% \begin{equation}
% \beta(X,\hat{Y}) = \frac{P_{D}(X,Y)}{P_{\bar{D}}(X,\hat{Y})} = \frac{P_{D}(Y|X)}{P_{\bar{D}}(\hat{Y}|X)}
% \label{eq:t-expected-loss}
% \end{equation}

\begin{equation}
\beta(X,\hat{Y}) = \frac{P_{D}(X,Y)}{P_{\bar{D}}(X,\hat{Y})} = \frac{P_{D}(Y|X)}{P_{\bar{D}}(\hat{Y}|X)}
\label{eq:t-revision-loss}
\end{equation}

With the true distribution $D$ unknown, we use the known transition matrix T under the assumption $P_{D}(X) = P_{\hat{D}}(X)$ to empirically estimate the true distribution $D$.

% \begin{equation}
% \widetilde{T} = \begin{bmatrix}
% 0.439 & 0.301 & 0.259\\
% 0.283 & 0.467 & 0.249\\
% 0.278 & 0.290 & 0.431
% \end{bmatrix}
% \caption{Estimated transition matrix for Cifar10 dataset}
% \end{equation}

\subsubsection{Formulation}

The Importance Re-weighting technique is used to rewrite the expected risk w.r.t clean data 
% (\ref{eq:ir-expected-risk}) 
given noise is independent of instances in a multi-class setting which also avoids the computationally intensive inverse of the transition matrix. The risk-consistent estimator (\ref{eq:ir-estimated-risk}) is derived given $f(X) = argmax_j\in\{1,\dots,C\} g_j(X)$ where $g_j(X)$ is an estimate for $P(Y=j|X)$ and $w$ denotes the loss function is weighted \cite{xia2019t_revision}. 

% \begin{equation}
% \mathbb{E}_{(X,Y)\tilde{}\bar{D}} [\bar{\ell}(f(X),Y)] = \int_{x} \sum_{i}P_{\bar{D}}(X=x,\bar{Y}=i) \frac{P_{D}(\bar{Y}=i|X=x)}{P_{\bar{D}}(\bar{Y}=i|X=x)}\ell(f(x), i)dx
% \label{eq:ir-expected-risk}
% \end{equation}

\begin{equation}
\bar{R}_{n,w}(T,f) = \frac{1}{n} \sum \frac{g_{\bar{Y_i}}(X_i)}{(T^{\intercal} g)_{\bar{Y_i}}(X_i)} \ell(f(X_i), \bar{Y_i})
\label{eq:ir-estimated-risk}
\end{equation}

\begin{equation}
\beta(X,Y) = \frac{g_{\bar{Y_i}}(X_i)}{(T^{\intercal} g)_{\bar{Y_i}}(X_i)}
\label{eq:ir-beta}
\end{equation}

\begin{figure}[h]
\centering
\includegraphics[width=9cm]{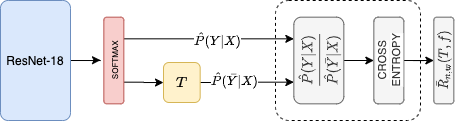}
\caption{Importance Re-weighting classifier.}
\label{fig:import-reweighting-loss-function}
\end{figure}

Using the provided $T$ as fixed, we use the softmax function to approximate $g(x)=\hat{P}(Y|X=x) \approx P(Y|X=x)$ and $T^{\intercal}g(x)=\hat{P}(\bar{Y}|X=x) \approx P(\bar{Y}|X=x)$. A ResNet-18 (\ref{fig:global_resnet}) deep learning architecture is used to learn $\beta(X,Y)$ which together with the cross-entropy loss function trains the classifier on the noisy dataset. Predictions on the clean dataset are produced by the classifier having learned $P(Y|X=x)$.

% \subsection{Transition Matrix Estimation}

\section{Noise Rate Estimation Method}\label{noise-rate-estimation}

We leverage the noisy dataset to learn the noisy class posteriors $P(\tilde{Y}|X)$ iteratively during the training of a deep learning classifier. The entire data set is used to identify instances which exhibit high noisy class posterior probabilities. We are able to estimate the fixed class-conditional noise empirically by estimating anchor points using learned information within the noisy dataset. The $T$ matrix (\ref{eq:t-matrix}) is then constructed comprising of a $C$ rows and columns, where $C$ represents the number of classes. Each entry indicates the likelihood of a noisy label given the true label. We developed a series of experiments to confirm the accuracy of this method in Section \ref{validating-our-estimator}.

\begin{equation}
T = \\
    \begin{bmatrix}
    P(\tilde{Y}=0|Y=0), P(\tilde{Y}=0|Y=1), P(\tilde{Y}=0|Y=2) \\
    P(\tilde{Y}=1|Y=0), P(\tilde{Y}=1|Y=1), P(\tilde{Y}=1|Y=2) \\
    P(\tilde{Y}=2|Y=0), P(\tilde{Y}=2|Y=1), P(\tilde{Y}=2|Y=2) \\
    \end{bmatrix}
\label{eq:t-matrix}
\end{equation}

\subsection{Formulation}

\begin{figure}[h]
    \centering
    \includegraphics[width=7cm]{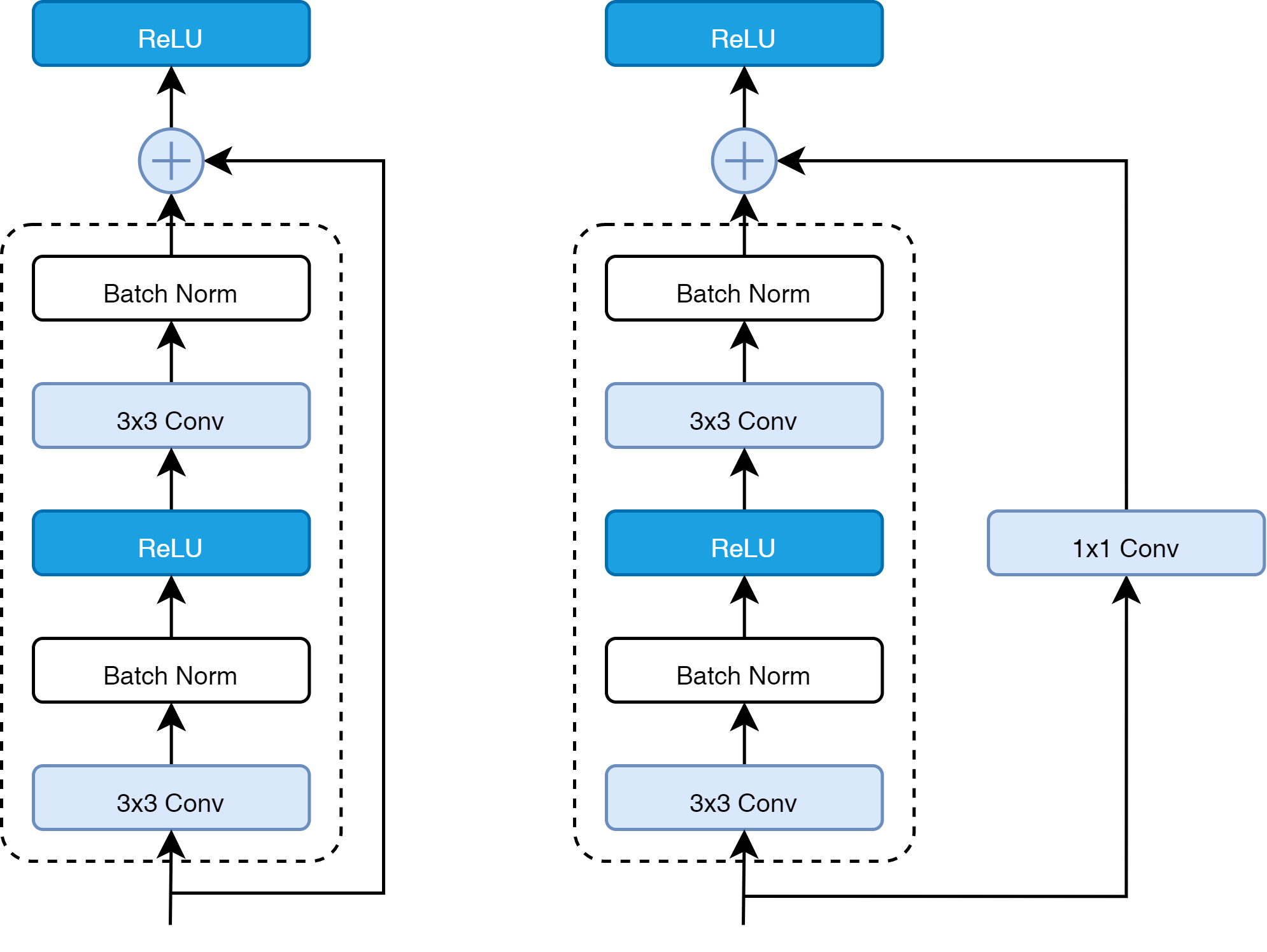}
    \caption{Basic ResNet Block without and with $1x1$ convolution.}
    \label{fig:block_resnet}
\end{figure}

\begin{figure}[h]
    \hspace*{-1.7cm}
    \includegraphics[width=17cm]{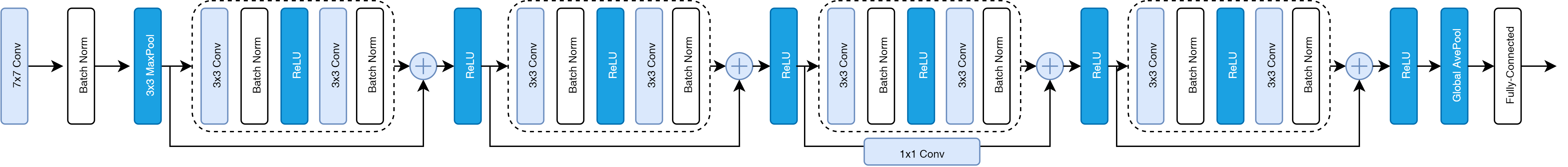}
    \caption{ResNet-18 Architecture.}
    \label{fig:global_resnet}
\end{figure}

We implemented the ResNet-18 which contains 18 layers grouped in 4 blocks. These blocks are composed of two convolutional layers, two batch normalization layers and a ReLU activation layer. The Figure \ref{fig:block_resnet} shows the architecture of the basic blocks and the Figure \ref{fig:global_resnet} presents the entire ResNet-18 architecture. Additionally, the softmax function is utilised to estimate the class of each instance.

% \begin{equation}
%     \hat{R}_{l,\bar{D}}(f) = \frac{1}{N}\sum_i^N l(f(x_i),\hat{y_{i}})
% \label{eq:t-estimation}
% \end{equation}

\section{Label Noise Methods with Unknown Flip Rates}

\subsection{T-Revision}

Using already established transition matrices, we analyse rigorous methods for multi-class image classification and compare each approach using top-1 accuracy. The T-revision approach suggested by \cite{xia2019t_revision} is investigated. In general, we initially configure the transition matrix by leveraging samples that are closer to the anchor points, including those with posterior probabilities of high approximate noisy class. By adding a slack vector afterwards, we alter the initial matrix, which will then be trained and validated along with the classifier by using only the noisy data.

The suggested method of T-revision works because by minimising the risk-consistent estimator, which is asymptotically equivalent to the anticipated risk w.r.t. clean data, we learn $\Delta T$ \cite{xia2019t_revision}. On the noisy validation set, the learned idle variable may also be validated, i.e. to verify whether $\widehat{P}( \overline{Y}|X = x )$ matches the validation set. The principle of this technique is close to that of the strategy of cross-validation. 

\subsubsection{Formulation}

\begin{figure}[h]
\centering
\includegraphics[width=9cm]{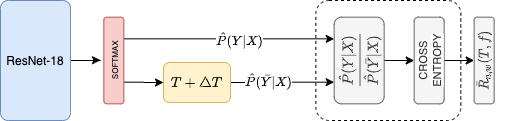}
\caption{T-Revision Re-weighting classifier.}
\label{fig:t-revision-reweighting-loss-function}
\end{figure}

The T-Revision approach is divided into two main stages as observed in Figure \ref{fig:t-revision-reweighting-loss-function}. We initially use the noisy training data to learn $\hat{T}$ by learning the noisy class posterior probabilities and approximating its initial state (\ref{noise-rate-estimation}). In a similar fashion as the Importance Re-weighting method (\ref{eq:ir-estimated-risk}) described previously, we now use $T$ and incrementally adjust it by $\triangle{T}$ throughout the training process whereas previously it was fixed.

% We decided to use ResNet as similar to this paper \cite{xia2019t_revision} did. 
% In the second stage, we just update the following function provided appropriate revision:

% \begin{equation}
% \bar{R}_{n, w}(T, f)=\frac{1}{n} \sum_{i=1}^{n} \frac{g_{\bar{Y}_{i}}\left(X_{i}\right)}{\left(T^{\top} g\right)_{Y_{i}}\left(X_{i}\right)} \ell\left(f\left(X_{i}\right), \bar{Y}_{i}\right)
% \end{equation}

% In which $f(X)=\arg \max _{j \in\{1, \ldots, C\}} g_{j}(X), g_{j}(X)$ is an approximation of $P(Y = j|X)$ and the subscript w implies the weighting of the loss feature. Note that $\bar{R}_{n, w}(T, f)$) has only one argument $g$ to learn if the true transition matrix $T$ is given.

% Detailed evidence is produced by the sophistication of the deep neural network hypothesis \cite{golowich2018size}.

\section{Experiment and Results} \label{experiment-and-results}
In this study, we create a series of experiments to prove the implementation of the previous methods and analyse the robustness to label noise of the different classifiers. Additionally, we validate the effectiveness of our transition matrix estimator using the true transition matrix provided.

The datasets used to conduct the experiments are FashionMNIST0.5, FashionMNIST0.6 and Cifar10. The FashionMNIST0.5 and FashionMNIST0.6 dataset contain 18000 images for training and validation. The images are in grayscale and the shape of each sample is $(28x28)$. On the other hand, the Cifar10 dataset contains 15000 colour images for training and validation, the shape of each sample is $(32x32x3)$. The classifiers will evaluate on the test set provided for each dataset, which each one contains 3000 samples.

\subsection{Pre-processing}

We apply no further pre-processing aside from a simple pixel value transformation whereby we convert 0-255 pixel values to values between 0-1 which make up the input features.

\subsection{Metrics}
To compare the performance and robustness of the different classifiers, we use the following metric:

\begin{itemize}
    \item \textbf{Top-1 Accuracy:} The output of each classifier will be evaluated using the top-1 precision metric. 
    
    $$
    \text { top-1 accuracy }=\frac{\text { number of correctly classified examples }}{\text { total number of test examples }}*100%
    $$
\end{itemize}

On the other hand, to evaluate the effectiveness of our transition matrix estimator we use:

\begin{itemize}
\label{ref:sum_avr}
    \item \textbf{Sum Average:} This metric allows us to measure the differences between the estimated transition matrix and the true transition matrix provided. 
    $$
    \text {sum average }=\frac{\sum_{i=1}^m\sum_{j=1}^n |\widehat{T}_{ij}|}{\sum_{i=1}^m\sum_{j=1}^n |T_{ij}|}
    $$
    
    where $T$ denotes the true transition matrix and $\widehat{T}$ denotes the estimated transition matrix.
\end{itemize}

To correctly perform the evaluation of these experiments, we train each classifier 10 times with the different training and validation sets generated by random sampling. Then we extract both the mean and the standard deviation of the accuracy of the test.

\subsection{Hardware}
The code has been developed using Python 3x and Jupyter Notebook and we have taken advantage of the computational capacity provided by Google Colab. However, a computer with the following specifications could easily run these experiments:

\begin{table}[h]
\centering
\begin{tabular}{|c|c|} 
\hline
Model & MSI GE63VR 7RF           \\ 
\hline
Processor & Intel Core i7 7700HQ 2.80GHz           \\ 
\hline
RAM & 16gb  \\ 
\hline
Graphics & GTX 1070 8GB  \\ 
\hline
\end{tabular}
\end{table}

\subsection{Transition Matrix Estimator}
In this section, we estimate the transition matrix for the Cifar10 dataset using the transition matrix estimator proposed in the previous section. Additionally, we evaluate the effectiveness of our estimator using the provided transition matrices for the FashionMNIST0.5 and FashionMNIST0.6 datasets.

\subsubsection{Estimate Transition Matrix - Cifar10}
As we mentioned previously, we used the ResNet architecture to learn the noisy class posterior $P(\widetilde{Y}|X)$ and empirically estimate the anchor points on the noisy dataset and use these anchor points to estimate the transition matrix. The Figure \ref{fig:block_resnet} shows the architecture of the basic blocks and the Figure \ref{fig:global_resnet} presents the entire ResNet-18 architecture.

The parameters used to train the neural network is contained in the Table \ref{tab:estimation_configuration}. In this case and to properly learn the noisy class posterior we trained the neural network during $10$ iterations using the Stochastic Gradient Descent (SGD) and Cross-Entropy as the loss function.

\begin{table}[h!]
\centering
    \begin{threeparttable}
    \begin{tabular}{|c|c|} 
    \hline
    \textbf{Iterations} & $10$\\ 
    \hline
    \textbf{Optimizer} & SGD\tnote{1}             \\ 
    \hline
    \textbf{Learning Rate} (\textit{lr}) & $0.001$  \\ 
    \hline
    \textbf{Momentum} & $0.9$ \\
    \hline
    \textbf{Loss Function} & Cross-Entropy  \\ 
    \hline
    \end{tabular}
    \begin{tablenotes}
        \item[1] Average Stochastic Gradient Descent.
    \end{tablenotes}
    \end{threeparttable}
\caption{Configuration used to train the ResNet-18 architecture to estimate the transition matrix.}
\label{tab:estimation_configuration}
\end{table}

The transition matrix $\widehat{T}$ estimated using the noisy class posterior for the Cifar10 dataset is presented below:

\begin{equation}
\widehat{T} = \begin{bmatrix}
0.439 & 0.301 & 0.259\\
0.283 & 0.467 & 0.249\\
0.278 & 0.290 & 0.431
\end{bmatrix}
\end{equation}

\subsubsection{Validating the effectiveness of our estimator} \label{validating-our-estimator}

To analyse the effectiveness of our estimator we use the provided transition matrices for the FashionMNIST0.5 and FashionMNIST0.6 datasets. We trained the ResNet-18 model using these datasets to learn the noisy class posterior and then generate the estimated transition matrix. Finally, we evaluate the difference between the transition matrix estimated by our estimator and the true matrix using the sum average metric (section \ref{ref:sum_avr}). The results are shown in the Table \ref{tab:comparison_t}.

We can observe how the estimated matrices $\widehat{T}$ using our estimator are very similar to the true matrices $T$ provided. For the FashionMNIST0.5 dataset, the sum average error between $\widehat{T}$ and $T$ is around $0.158$. On the other hand, the error for the estimated transition matrix using the FashionMNIST0.6 dataset is around $0.09$. In this case, the estimator manages to approximate the matrix very well and the small error may be due to the number of decimals that the estimated matrix contains. We can conclude that our estimator produces a very satisfactory result and we can consider the estimated transition matrix for the Cifar10 dataset accurate.

\begin{table}[h]
\centering
\begin{tabular}{cccc} 
        \toprule
        Dataset & $T$ & $\widehat{T}$ & Sum Average Error\\
        \midrule
        Cifar10 & - & $\begin{bmatrix}0.439 & 0.301 & 0.259\\0.283 & 0.467 & 0.249\\0.278 & 0.290 & 0.431\end{bmatrix}$ & -\\
        \bottomrule
        FashionMNIST0.5 & $\begin{bmatrix}0.5 & 0.2 & 0.3\\0.3 & 0.5 & 0.2\\0.2 & 0.3 & 0.5\end{bmatrix}$ & $\begin{bmatrix}0.545 & 0.224 & 0.229\\0.231 & 0.488 & 0.280\\0.285 & 0.213 & 0.501\end{bmatrix}$ & \textbf{0.158}\\
        \bottomrule
        FashionMNIST0.6 & $\begin{bmatrix}0.4 & 0.3 & 0.3\\0.3 & 0.4 & 0.3\\0.3 & 0.3 & 0.4\end{bmatrix}$ & $\begin{bmatrix}0.475 & 0.250 & 0.274\\0.273 & 0.433 & 0.292\\0.289 & 0.281 & 0.429\end{bmatrix}$ & \textbf{0.09}\\
        \bottomrule
\end{tabular}
\caption{Comparison between the provided true transition matrix and the estimated transitions matrix using our estimator.}
\label{tab:comparison_t}
\end{table}

\subsection{Robustness to Noisy Labels}
In this section, we provide the results for the analysis of the robustness to noisy labels of the classifiers mentioned previously. 

\subsubsection{Forward Learning}
\begin{figure}[h]
\centering
\includegraphics[width=13cm]{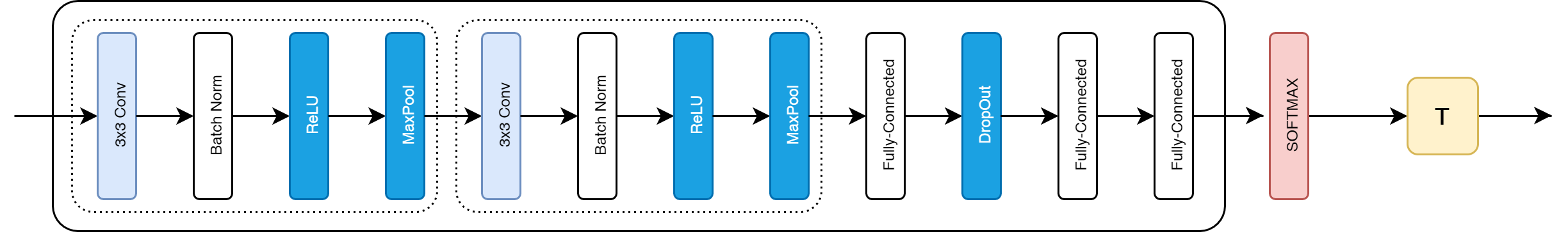}
\caption{Convolutional Neural Network implemented with Forward Learning.}
\label{fig:global_cnn}
\end{figure}

As we mentioned previously, we implemented a Convolutional Neural Network with the Forward Learning method. The architecture of this neural network is presented in Figure \ref{fig:global_cnn}.

\begin{figure}[h]
\centering
\includegraphics[width=9cm]{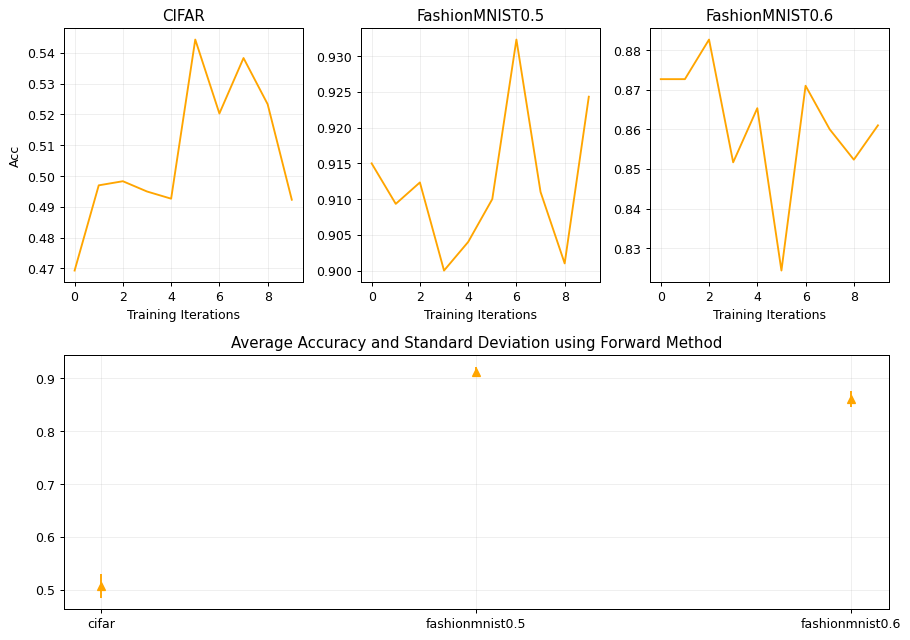}
\caption{Average Accuracy and Standard Deviation using Forward Method.}
\label{fig:forward_test}
\end{figure}

In Figure \ref{fig:forward_test} we present the accuracy on the test set using Forward Learning for the different datasets. The top panel of the figure contains the accuracy for each training time and the bottom panel shows the mean and standard deviation value.

It can be observed how the performance for the FashionMNIST0.5 and FashionMNIST0.6 is pretty similar, around $0.87$. On the other hand, the accuracy of the Cifar10 dataset is around $0.5$.

\subsubsection{Importance Re-weighting}
In this experiment, we trained the ResNet model defined previously and we used importance re-weighting to adjust the effects of noisy labels. The architecture of the neural network is presented in Figure \ref{fig:global_resnet}.

\begin{figure}[h]
\centering
\includegraphics[width=9cm]{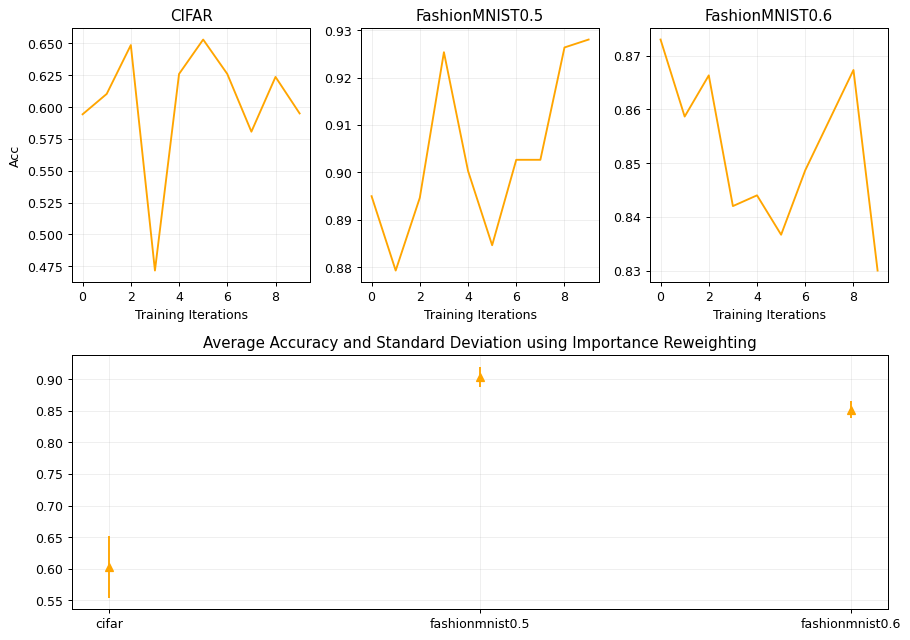}
\caption{Average Accuracy and Standard Deviation using Importance Re-weighting.}
\label{fig:impreweighting_test}
\end{figure}

Figure \ref{fig:impreweighting_test} presents the accuracy on the test set using the importance re-weighting method. As we mentioned previously, the top panel shows the evolution of the accuracy for each training time and the bottom panel contains the average accuracy and the standard deviation for each dataset. The dataset Cifar10 has experienced an increase in accuracy from $0.50$ to $0.60$. However, the accuracy for the FashionMNIST0.5 and FashionMNIST0.6 remains practically the same, around $0.90$ and $0.85$ respectively.

\subsubsection{T-Revision}
As we mentioned previously, we implemented T-Revision using the ResNet model defined in the last section.

\begin{figure}[h]
\centering
\includegraphics[width=9cm]{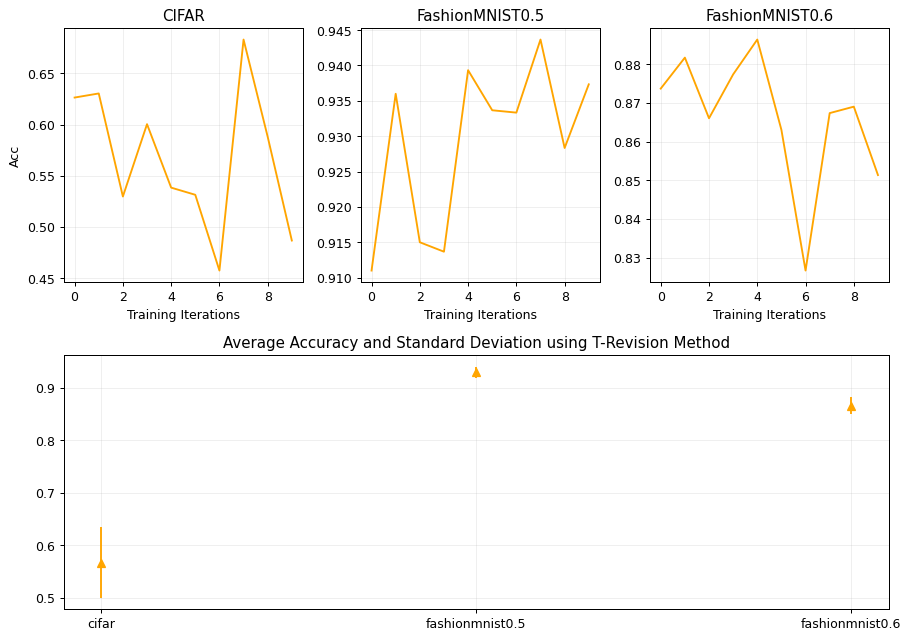}
\caption{Average Accuracy and Standard Deviation using T-Revision.}
\label{fig:trevision_test}
\end{figure}

The results presented in Figure \ref{fig:trevision_test} show how the accuracy for the dataset Cifar10 experienced a decrease from $0.60$ to $0.58$ with respect to the importance re-weighting method. Nevertheless, the accuracy for the FashionMNIST0.5 and FashionMNIST0.6 dataset has slightly increased from $0.90$ to $0.92$ and from $0.85$ to $0.88$ respectively. 

As T-Revision allows us to learn $\Delta T$, in the Table \ref{tab:corr_t} we present the value of $\Delta T$, $\widehat{T} + \Delta T$ for each dataset. 

\begin{table}[h]
\hspace*{-1.2cm}
%\centering
    \begin{threeparttable}
    \begin{tabular}{cccc} 
        \toprule
        Dataset & $\widehat{T}$ & $\Delta T$ & $(\widehat{T} + \Delta T)$\\
        \midrule
        Cifar10\tnote{1} & $\begin{bmatrix}0.439 & 0.301 & 0.259\\0.283 & 0.467 & 0.249\\0.278 & 0.290 & 0.431\end{bmatrix}$ & $\begin{bmatrix}0.0332 & 0.0366 & 0.0286\\0.0416 & 0.0449 & 0.0462\\0.0322 & 0.0508 & 0.0372\end{bmatrix}$ & $\begin{bmatrix}0.4726 & 0.3386 & 0.2872\\0.3246 & 0.5122 & 0.2960\\0.3111 & 0.3409 & 0.4683\end{bmatrix}$\\
        \bottomrule
        FashionMNIST0.5 & $\begin{bmatrix}0.5 & 0.2 & 0.3\\0.3 & 0.5 & 0.2\\0.2 & 0.3 & 0.5\end{bmatrix}$ & $\begin{bmatrix}0.0279 & 0.0216 & 0.0400\\0.0243 & 0.0219 & 0.0228\\0.0307 & 0.0331 & 0.0282\end{bmatrix}$ & $\begin{bmatrix}0.5279 & 0.2216 & 0.3400\\0.3243 & 0.5219 & 0.2228\\0.2307 & 0.3331 & 0.5282\end{bmatrix}$\\
        \bottomrule
        FashionMNIST0.6 & $\begin{bmatrix}0.4 & 0.3 & 0.3\\0.3 & 0.4 & 0.3\\0.3 & 0.3 & 0.4\end{bmatrix}$ & $\begin{bmatrix}0.0482 & 0.0340 & 0.0452\\0.0389 & 0.0447 & 0.0420\\0.0463 & 0.0338 & 0.0434\end{bmatrix}$ & $\begin{bmatrix}0.4482 & 0.3340 & 0.3452\\0.3389 & 0.4447 & 0.3420\\0.3463 & 0.3338 & 0.4434\end{bmatrix}$\\
        \bottomrule
    \end{tabular}
    \begin{tablenotes}
        \item[1] For the Cifar10 dataset, the $\widehat{T}$ was estimated using the estimator implemented in the previous section.
    \end{tablenotes}
    \end{threeparttable}
    \caption{Transition matrices with the correction $\Delta T$ learned using T-Revision.}
    \label{tab:corr_t}
\end{table}

\subsubsection{Comparison}
This section contains a comparison between the results previously presented and the baseline model for each dataset. The baseline model refers to the same model (ResNet or CNN) trained without applying any technique to correct the effect of the noisy labels. We analyse if the methods implemented to improve the performance of the baseline model.

\begin{figure}[h]
\centering
\includegraphics[width=9cm]{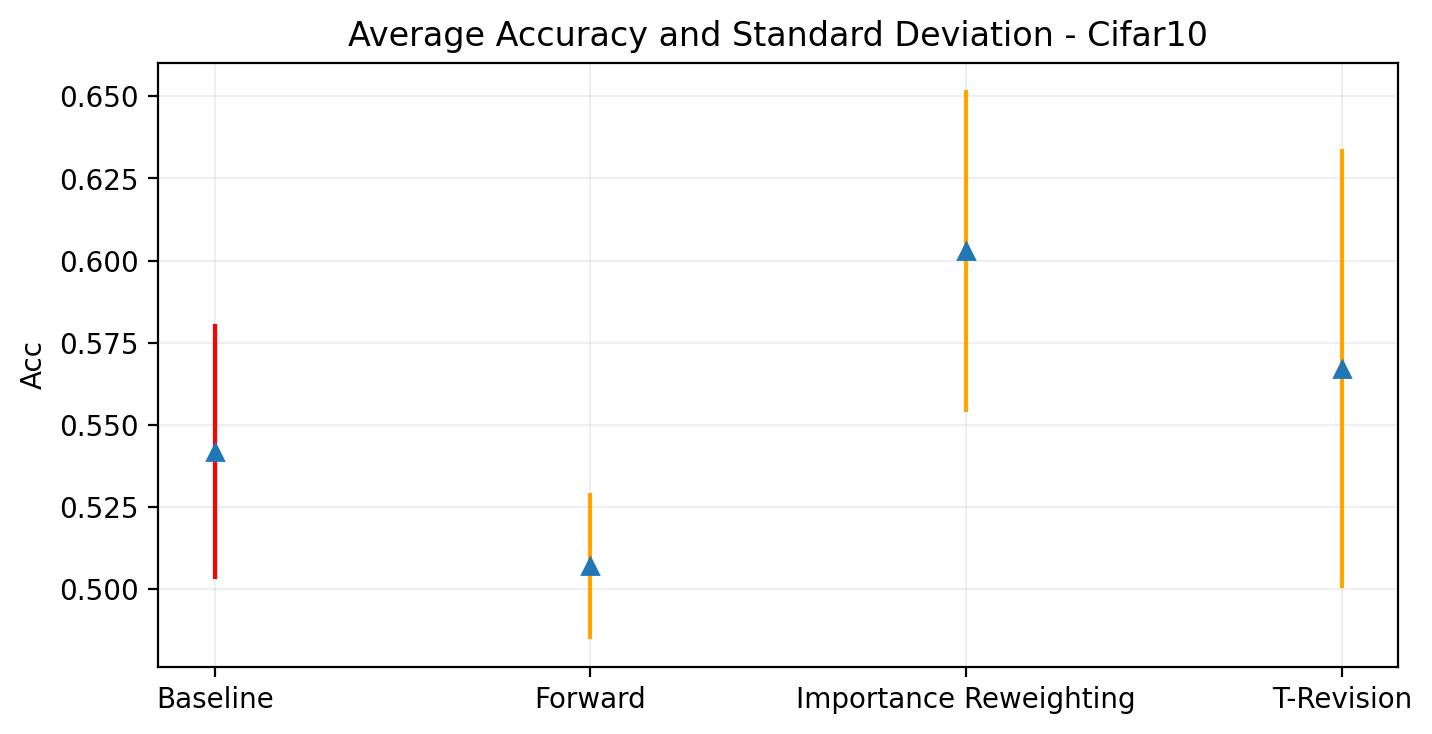}
\caption{Comparison of the average accuracy and standard deviation for different methods - Cifar10}
\label{fig:baseline_cifar10}
\end{figure}

{\textbf{Cifar10}}\\
Figure \ref{fig:baseline_cifar10} shows the average accuracy and standard deviation for the Forward, Importance Re-weighting and T-Revision methods. As we can see, the accuracy using the Forward method is the lowest of all, around $0.51$. On the other hand, it seems that the performance of the baseline model is better than the Forward method, around $0.54$. However, the accuracy obtained using Importance Re-weighting and T-Revision is slightly higher with respect to the baseline model, around $0.60$ and $0.565$ respectively.

\begin{figure}[h]
\centering
\includegraphics[width=9cm]{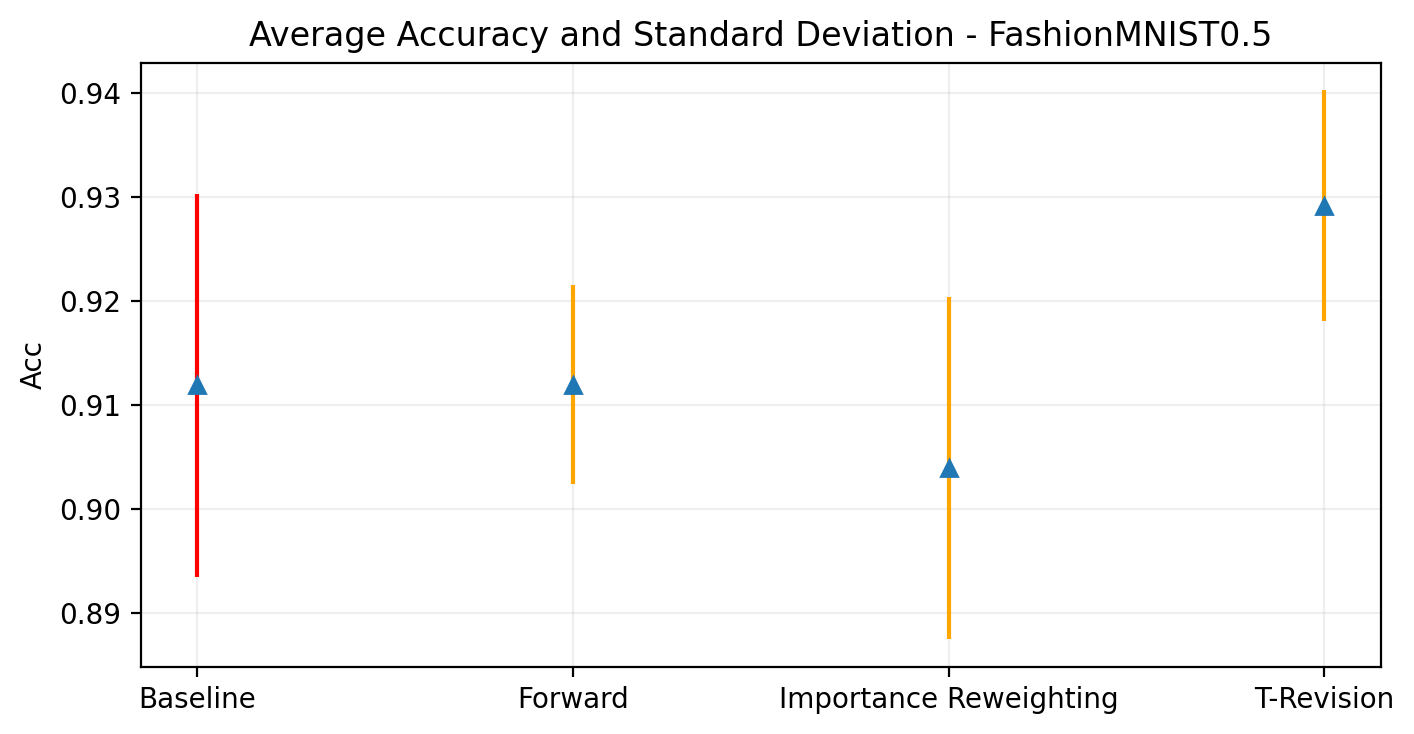}
\caption{Comparison of the average accuracy and standard deviation for different methods - FashionMNIST0.5}
\label{fig:baseline_fashion05}
\end{figure}

{\textbf{FashionMNIST0.5}}\\
As we can observe in the Figure \ref{fig:baseline_fashion05}, the accuracy obtained for FashionMNIST0.5 is pretty higher and it seems that we achieved a very good result just using the baseline model, around $0.91$. Additionally, the accuracy using the Forward and Importance Re-weighting method is very similar to the one obtained using the baseline model, around $0.91$ and $0.905$ respectively. On the other hand, using T-Revision we observe an increase in accuracy up to $0.93$.

As we mentioned before, the performance using the baseline model, T-Revision and Importance Re-weighting method is very similar and we can therefore draw the conclusion that the transition matrix provided is not very accurate and thus the result of these methods is highly similar to the baseline model. However, the T-Revision method allows us to slightly increase the accuracy, this is because this method allows us to update the transition matrix during the training and thus obtain a more accurate matrix. 

\begin{figure}[h]
\centering
\includegraphics[width=9cm]{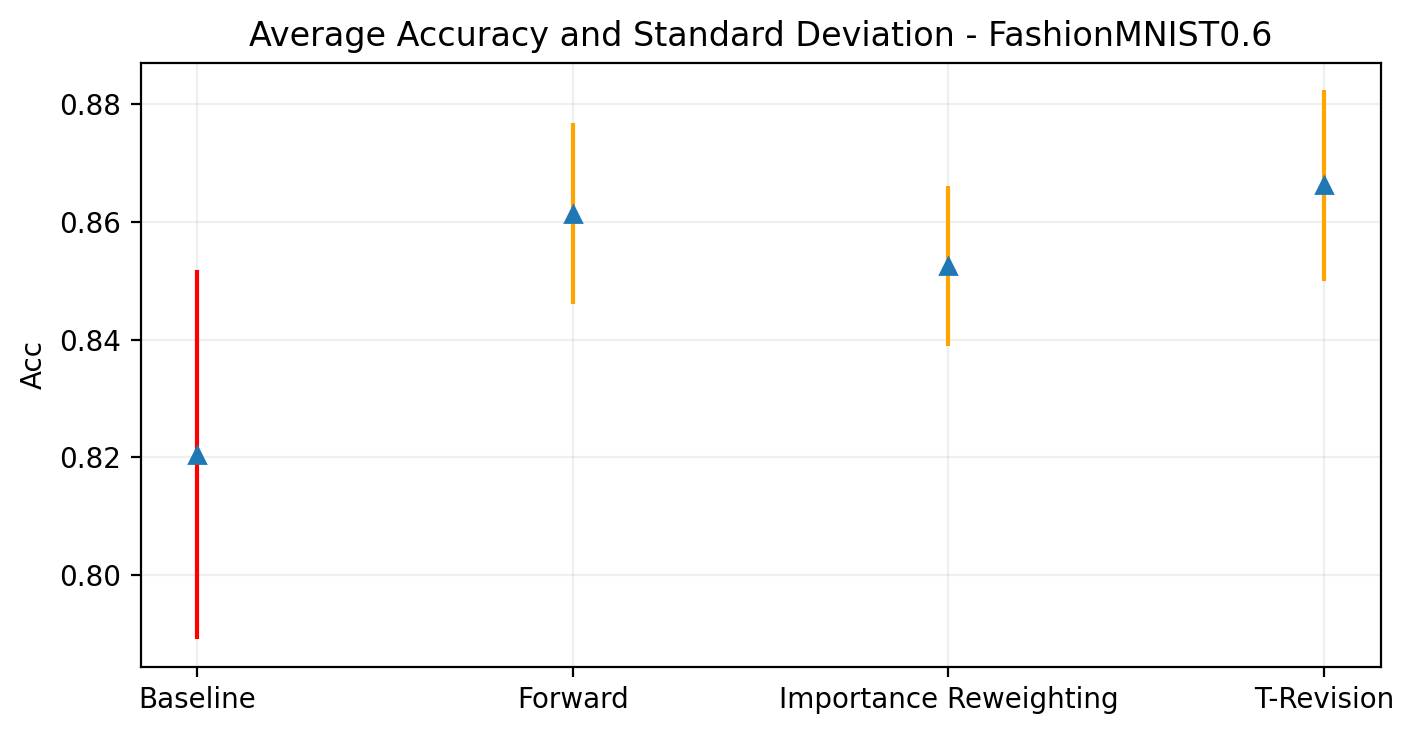}
\caption{Comparison of the average accuracy and standard deviation for different methods - FashionMNIST0.6}
\label{fig:baseline_fashion06}
\end{figure}

{\textbf{FashionMNIST0.6}}\\
Figure \ref{fig:baseline_fashion06} shows that in contrast to the results obtained in the other datasets, we obtain higher accuracy with respect to the baseline model in all the methods used. In this case, the baseline model obtained an accuracy of around $0.82$. Nevertheless, the accuracy achieved with Forward and Importance Re-weighting method is very similar, around $0.86$. On the other hand, T-Revision is the method which obtained the highest accuracy, about $0.865$.

In this case and observing the results obtained, we can conclude that the transition matrix provided is accurate and that the transition matrix for the cifar10 dataset has been accurately estimated. As we mentioned before, the T-Revision method allows us to obtain a higher accuracy due to update the transition matrix and this enables us to get a more accurate result.

\section{Conclusion}
In this project, we presented and implemented the formulation of three different methods to make the models robust to noisy labels. Additionally, we implemented a transition matrix estimator to estimate the transition matrix for the Cifar10 dataset. We analysed the performance of these methods and the effectiveness of our estimator.

It could be observed that the estimator implemented to estimate the transition matrix produces a highly accurate result. We evaluated its effectiveness using the transition matrices provided for FashionMNIST0.5 and FashionMNIST0.6 dataset and the errors obtained from comparing the matrices obtained using our estimator and the true matrices are very small so we concluded that our estimator produces a very accurate result. 

On the other hand, we analysed the behaviour of our classifiers to noisy labels when the Forward, Importance Re-weighting and T-Revision methods are applied. Firstly, we concluded that Importance Re-weighting is the method which achieved the highest accuracy for the Cifar10 dataset, followed closely by T-Revision. In parallel, we found that the transition matrix provided for the FashionMNIST0.5 dataset may not be very accurate because the Forward and Importance Re-weighting methods obtain a very similar result to the baseline model. However, the T-Revision achieved a higher accuracy thus, we concluded that the transition matrix for this dataset may not be accurate and we should use T-Revision or our estimator to obtain a better approximation of the transition matrix for this case. Finally, we observed how the models trained using the FashionMNIST0.6 dataset experienced an improvement using the Forward, Importance Re-weighting or T-Revision methods. 

In the future, we can extend this study in the following aspects:

\begin{itemize}
    \item Study if it is possible to estimate the transition matrices at the same time as we train the classifiers.
    \item Extend this work to more complex datasets, such as ImageNet and analyse its performance.
    \item Estimate the transition matrix for the FashionMNIST0.5 dataset and analyse if the provided transition matrix has been well estimated.
\end{itemize}

%Future studies will boost the estimation process by adding noise structure priors, assuming, for example, a low T-rank. Improvements in this direction can also expand the applicability of situations that are significantly multi-class. Whether example-dependent noise can be used in our approach \cite{42xiao2015learning, 25menon2016learning} remains an open question. Finally, in the framework of \cite{17krause2016unreasonable}, we foresee the use of our methodology as a tool for pre-training models with noisy data from the Internet.

\newpage

\printbibliography

\newpage
\section{Appendix}

\href{https://github.com/alejandrods/Analysis-of-classifiers-robust-to-noisy-labels}{\textit{Click here to access to the repository with the code.}}{}\\

The code of this work is provided in the Jupyter Notebook called \emph{Analysis\_of\_classifiers\_robust\_to\_noisy\_labels.ipynb} and the experiments were executed using Python 3x in Google Colab. 

\begin{figure}[h]
\centering
\includegraphics[width=12cm]{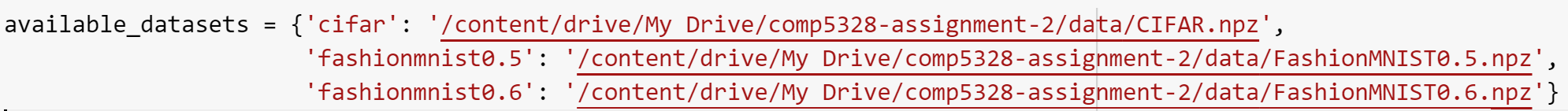}
\centering
\caption{Provide a path to the target dataset.}
\label{fig:drive_1}
\end{figure}

To execute the code you just need to go through the notebook and executing all cells. The first section called \emph{Load and preprocessing data} contains the function to load and preprocess the images. In this section, you should add the path to the folder with the datasets, in case you use Google Colab you need to add the datasets to your Drive account and mount your Drive folder into the notebook (figure \ref{fig:drive_1}).

Section \emph{Utils Functions} contains functions that are used in the implementation of our algorithms. The PyTorch function to load the dataset, generate predictions as well as the function to estimate the transition matrix are contained in this section. Furthermore, the section called \emph{Classifiers robust to label noise} contains the implementation of the different proposed methods. Finally, section \emph{Experiments} contains the experiments we have executed to analyse the effectiveness of our estimator and the performance of the classifiers to noisy labels using the different methods.

\clearpage
\vskip 1in

\end{document}